\documentclass[journal]{IEEEtran}
\usepackage{graphicx}
\usepackage{amssymb}
\usepackage{amsmath}
\usepackage{amsthm}
\usepackage{hyperref}
\usepackage{siunitx}
\usepackage{booktabs}
\usepackage{multirow}
\usepackage{bbm}
\usepackage{xcolor}
\usepackage{dblfloatfix}
\usepackage{authblk}
\usepackage{multicol}
\usepackage{siunitx}

\newtheorem*{example*}{Example}

\newcommand{\x}{\boldsymbol{x}}
\newcommand{\ba}{\boldsymbol{a}}

\begin{document}

\title{Toward Explainable AI for Regression Models}

\author[1]{Simon Letzgus$^{\dag}$\thanks{$^\dag$ S.\ Letzgus and P.\ Wagner contributed equally to this work.}}
\author[2]{Patrick Wagner$^{\dag}$}
\author[1]{Jonas Lederer}
\author[2,3]{Wojciech Samek$^*$\thanks{$^*$ Corresponding authors: W.\ Samek, K.-R.\ M\"uller and G.\ Montavon.}}
\author[1,3,4,5,6]{Klaus-Robert Müller$^*$}
\author[1,3]{Gr\'egoire Montavon$^*$}

\affil[1]{Machine Learning Group, Technische Universität Berlin, 10587 Berlin,
Germany}
\affil[2]{Department of Artificial Intelligence, Fraunhofer Heinrich Hertz Institute, 10587 Berlin, Germany}
\affil[3]{BIFOLD -- Berlin Institute for the Foundations of Learning and Data, 10587 Berlin, Germany}
\affil[4]{Department of Artificial Intelligence, Korea University, Seoul 136-713, South Korea}
\affil[5]{Max Planck Institute for Informatics, Stuhlsatzenhausweg 4, 66123 Saarbrücken,
Germany}
\affil[6]{Google Research, Brain Team, Berlin,
Germany}

\maketitle

\begin{abstract}
In addition to the impressive predictive power of machine learning (ML) models, more recently, explanation methods have emerged that enable an interpretation of complex non-linear learning models such as deep neural networks. Gaining a better understanding is especially important e.g. for safety-critical ML applications or medical diagnostics etc. While such Explainable AI (XAI) techniques have reached significant popularity for classifiers, so far little attention has been devoted to XAI for regression models (XAIR). In this review, we clarify the fundamental conceptual differences of XAI for regression and classification tasks, establish novel theoretical insights and analysis for XAIR, provide demonstrations of XAIR on genuine practical regression problems, and finally discuss the challenges remaining for the field.
\end{abstract}

\begin{IEEEkeywords}
Explainable AI, Regression, Deep Neural Networks.
\end{IEEEkeywords}

\IEEEpeerreviewmaketitle

\section{Introduction}

Machine learning, in particular deep learning, has supplied a vast number of scientific and industrial applications with powerful predictive models. As ML models are being increasingly considered for high-stakes autonomous decisions, there has been a growing need for gaining trust in the model without giving up predictive power. Explainable artificial intelligence (XAI) has developed as a response to the need of validating these highly powerful models \cite{DBLP:series/lncs/11700,DBLP:journals/inffus/ArrietaRSBTBGGM20,DBLP:journals/pieee/SamekMLAM21}. Taking ML and XAI together, these technologies also offer a way of gaining new insights, e.g.\ nonlinear relations, into the complex data generating processes under study.

So far, the main body of work in the field of XAI has revolved around explaining decisions of classification models, often in the context of image recognition tasks \cite{baehrens2010explain,DBLP:conf/eccv/ZeilerF14,bach-plos15,DBLP:conf/cvpr/ZhouKLOT16,DBLP:conf/kdd/Ribeiro0G16}. Regression, a major workhorse in ML and signal processing, has essentially only received little attention. In practice to date, XAI approaches designed for classification are applied for regression problems. While such naive application to regression can occasionally still yield useful results, we will show in this paper that appropriate theoretically well-founded explanation models are necessary and overdue. For example, when explaining classification models, we can rely on the implicit knowledge associated with the class, and assume a decision boundary between the two or more classes. Additionally, the output itself can conveniently serve as a measure of model uncertainty or even evidence against the respective class can be analysed. In regression, on the other hand, we find none of these beneficial properties while we aim to explain a highly aggregated and application-specific model output that often corresponds to a physical entity with an attached unit.

In this paper, we will outline multiple challenges that emerge when explaining regression models, and we show how popular methods such as Layer-Wise Relevance Propagation (LRP) \cite{bach-plos15}, Integrated Gradients \cite{DBLP:conf/icml/SundararajanTY17}, or the Shapley Value \cite{Shapley1953,DBLP:journals/jmlr/StrumbeljK10,DBLP:conf/nips/LundbergL17}, can be applied or extended in a theoretically well-founded manner to properly address them. Our efforts are guided by how to formulate the question for which we would like an explanation in a way that addresses the user's interpretation needs. In particular, we aim for an explanation that inherits the unit of measurement of the prediction (e.g.\ physical or monetary unit). The explanation should also be sufficiently contextualized, not only by being specific to each data point, but also by localizing the explanation around a relevant range of predicted outputs. As an example, LRP and many other explanation methods assume (sometimes implicitly) a zero-reference value relative to which they explain or expand \cite{DBLP:journals/pieee/SamekMLAM21}. While in the classification setting one naturally explains relative to the decision boundary, i.e., $f(\boldsymbol{x})=0$, in regression settings we consider the reference value to be a crucial parameter to integrate the desired context into the explanation and avoid the latter to be dominated by (uninteresting) coarse effects. Therefore, we generally discourage practitioners to apply XAI methods developed for classification problems in an out-of-the-box manner to regression models.

We will provide several illustrative regression examples from various fields of signal processing, motivating the need for a distinct treatment of the regression and classification explanation problems respectively. In particular, we showcase our approach on a large CNN model for age prediction in face images, and we also demonstrate the capability of XAI for regression to deliver targeted scientific insights into the energy structure of molecules. Moreover, we have established \url{www.xai-regression.org} as a point of contact for the community with a collection of related papers and applications, and complementary code can be found on \url{https://github.com/sltzgs/xai-regression}.

\section{A Brief Review of XAI}
\label{section:review}

The field of Explainable AI is very broad, as it must consider at the same time various types of ML models and many interpretability requirements. The resulting multitude of methods can be divided along various conceptual lines. One can differentiate, for example, between the explanation of a model's overall (global) decision strategy \cite{DBLP:conf/nips/NguyenDYBC16,LapNCOMM19, simonyan2014deep} and uncovering a model's reasoning related to a specific sample (local) \cite{bach-plos15,DBLP:conf/kdd/Ribeiro0G16,baehrens2010explain}. Methods can also be distinguished by how they present an explanation to the user: Some methods generate a maximally activating pattern \cite{simonyan2014deep}, which can be interpreted to be prototypical for the decision strategy; some methods extract a contrastive example (criticism \cite{NIPS2016_5680522b}, counterfactuals \cite{wachter2018a}) which resembles the current data point, but without the features that cause a particular decision behavior; other methods identify a subset of features that are relevant to explain the decision outcome \cite{DBLP:conf/icml/ChenSWJ18}; and finally, methods commonly referred to as attribution, assign to each input feature a score representing the contribution of that feature to the model output $f(\x)$ \cite{DBLP:journals/jmlr/StrumbeljK10,bach-plos15,DBLP:conf/icml/SundararajanTY17}. A further distinction can be made between methods that explain based on individual input features \cite{DBLP:journals/jmlr/StrumbeljK10,baehrens2010explain,bach-plos15,DBLP:conf/icml/SundararajanTY17}, combinations of input features (e.g.\ \cite{caruana15,schnake2020xai,Eberle-BiLRP}), or higher-level concepts \cite{kim2018interpretability}. While explanation methods are often evaluated based on their technical merit (e.g.\ accuracy, runtime) \cite{SamTNNLS17}, an increasingly relevant question is whether these explanations enable the human to truly understand the model at hand (also known as causability \cite{HOLZINGER202128}) and whether these explanations can be turned into meaningful insights and decisions \cite{DBLP:journals/corr/Doshi-VelezK17,DBLP:journals/access/RoscherBDG20,binder2021morphological,unke2021spookynet}.

\smallskip

This contribution will focus on single-instance (local), attribution-based explanations that assign a share of the model output $f(\x)$ to the individual features of the respective input sample $\x$. While the focus on attribution may appear somewhat narrow, we note that attribution can serve as a building block for a broader class of explanations. For example, the SpRAy method \cite{LapNCOMM19} enables to turn a large collection of single-instance explanations into a single concise dataset-wide explanation. Also, attributions can be easily turned into a set of relevant features, e.g.\ by thresholding attribution scores. Lastly, because attributions characterize the decision function at a given data point \cite{SamTNNLS17}, one can synthesize prototypical patterns or counterfactual examples from them as well, by removing the least or most relevant features. For the purpose of introducing the concept of attribution, it is useful to start with a simple setting such as a linear classifier. A linear (binary) classifier is typically implemented by a function
\begin{align}
f(\x) = \sum_{i=1}^d w_i x_i + b
\label{eq:linear}
\end{align}
where $\x = (x_1,\dots,x_d)$ is the vector of input features, and where $w_1,\dots,w_d,b$ are the parameters of the model. This function is typically followed by a sign function or some sigmoid function, where the output of such function either predicts the class directly, or assigns a probability of membership of the data point to the given class. The quantity $f(\x)$ can be interpreted as the evidence for\,/\,against a particular class. In practice, it is convenient to focus on explaining the latter rather than the actual probability value or the classification result \cite{montavon2018methods}.

\medskip

To obtain meaningful attributions, one also often considers the task of explanation relative to some neutral \textit{reference} point or counterfactual, which could be for example the origin in input space, or a point similar to $\x$ but without the pattern that causes the particular classification outcome \cite{bach-plos15,DBLP:conf/iccv/FongV17,wachter2018a}. A reference point can be for example a root point $\widetilde{\x}$ of the prediction function $f$ (i.e.\ $f(\widetilde{\x}) = 0$) with low distance $\|\x-\widetilde{\x}\|$.

\medskip

The concept of attribution and neutral point can be nicely illustrated for the simple linear example above. Assume we have computed some root point $\widetilde{\x}$ point of the function $f$. We can show that the linear model in Eq.\ \eqref{eq:linear} can be rewritten as:
\begin{align}
f(\x) &= \sum_{i=1}^d w_i \cdot (x_i - \widetilde{x}_i)
\label{eq:taylor}
\end{align}
where the root point now appears explicitly. With such reformulation, the bias term vanishes and the function is now structured as a simple sum over input features, which can also be interpreted as a first-order Taylor expansion at root point $\widetilde{\x}$ \cite{bach-plos15}. Equation \eqref{eq:taylor} gives rise to a natural attribution scheme, which is to score input features according to the elements of the sum, i.e.
\begin{align}
R_i = w_i \cdot (x_i - \widetilde{x}_i).
\label{eq:attr-linear}
\end{align}
The score $R_i$ can be interpreted as the contribution of input feature $i$ to the function output. Each contribution is the product of the feature magnitude itself relative to the reference point ($x_i-\widetilde{x}_i$) and the sensitivity of the model output to that feature ($w_i$). The vector $\boldsymbol{R} = (R_1,\dots,R_d)$ forms the {\it explanation}. Note that this explanation method differs from a sensitivity analysis, where only the parameters $w_i$ would be involved and where the same explanation would be consequently obtained for every input vector. Here instead, each input vector $\x$ produces an individual explanation.

\medskip

In most practical applications, the ML models are nonlinear, which renders the simple method above inapplicable. While the question of explaining nonlinear models is still an active research topic, several approaches have been developed, with different assumptions on the model structure, the level of access we have to the structure of the model, model accuracy requirements, explanation accuracy requirements, and the amount of available compute power \cite{DBLP:journals/jmlr/StrumbeljK10,bach-plos15,DBLP:conf/kdd/Ribeiro0G16,DBLP:conf/icml/SundararajanTY17,DBLP:conf/iclr/BrendelB19}.

In this paper, we place our focus on `post-hoc' explanations where we assume a given and already trained model (usually represented by some function $f$) and try to attribute the prediction for each data sample to its input features in a meaningful manner. In comparison to the `ante-hoc' setting where we can set the structure of the model before training in a manner that makes it easy to extract an explanation, the post-hoc setting makes no such assumption. Instead, the post-hoc setting decouples the task of model building and the task of explanation, and it simply assumes that one has chosen the most suitable or best-performing model for the given task (e.g.\ the one delivering the highest accuracy, or incorporating the desired invariances). Within the scope of post-hoc attribution methods that we adopt here, three families of methods can be distinguished, (1) feature removal, (2) gradient-based, and (3) backward propagation. We present below these families of methods, highlighting some of their respective members, which we will show later on to be particularly suitable to adapt from classification to regression.

\medskip

\subsection{Removal-based explanations}
\label{section:methods-removal}

Perhaps the most straightforward way of testing the contribution of a feature to the output of an ML model is to remove it and measure the difference at the output of the model. \textit{Shapley values} \cite{Shapley1953, DBLP:journals/jmlr/StrumbeljK10, DBLP:conf/nips/LundbergL17} provides a theoretical framework for this type of explanation. The framework originated in game theory where a related problem, that of sharing the total gain between a set of cooperating players, was considered. It was shown that given a fairly limited set of axioms that an explanation should satisfy, known as efficiency ($\sum_i R_i = f(\x)$, also known as conservation or completeness), symmetry, linearity, and null-player, the Shapley value \cite{Shapley1953} is one and the unique solution satisfying all axioms.

Let $\x$ be a data point composed of $N$ features. Let $\sum_{\mathcal{S}|i \notin \mathcal{S}}$ be a sum over all subsets of features that do not contain feature $i$, and $\x_\mathcal{S}$ the data point $\x$ where only features in $\mathcal{S}$ have been retained (the other features have been set to zero or the value of some meaningful reference point $\widetilde{\x}$). The Shapley value is given by:
$$
R_i = \sum_{\mathcal{S}|i \notin \mathcal{S}} \alpha_\mathcal{S} \cdot \big[f(\x_{\mathcal{S} \cup \{i\}}) - f(\x_{\mathcal{S}})\big]
$$
where $\alpha_\mathcal{S} = |\mathcal{S}|!(N-|\mathcal{S}|-1)/N!$. In other words, the Shapley value tests the effect of adding the feature $i$ assuming various subsets of features are present, and weighting the different subsets by the factor $\alpha_\mathcal{S}$.

The Shapley value can be applied to any function $f(\x)$, whether it is a neural network, a random forest, a kernel classifier, etc. Because there are exponentially many subsets that need to be evaluated, the Shapley value is however computationally infeasible for most problems with more than 15 or 20 input dimensions. Only for certain classes of models or allowing certain approximations, Shapley values can be computed in polynomial time \cite{lundberg2020local,DBLP:conf/icml/AnconaOG19}. For general models, random approximations of Shapley values where elements of the sum are sampled according to $\alpha_\mathcal{S}$ can be used, which allows scaling the analysis to higher dimensions. For a recent review of feature removal-based methods, we refer to \cite{Covert2021}.

\subsection{Gradient-Based explanations}
\label{section:methods-gradient}

Another set of methods bypasses the need to evaluate the function for multiple perturbations by relying instead on the gradient (e.g.~\cite{baehrens2010explain}). The latter can be extracted quickly with a single forward/backward pass in the function's computational graph. A simple approach that generalizes Eq.\ \eqref{eq:attr-linear} consists of replacing the weight $w_i$ by the gradient evaluated locally and integrating it locally along some trajectory $\{\boldsymbol{\x}(t) \colon 0 \leq t \leq 1\}$ connecting the root point and the data point i.e.\
$$
R_i = \int_{0}^1 \frac{\partial f}{\partial x_i(t)} \cdot \frac{\partial x_i(t)}{\partial t} \cdot dt
$$
The method is known as \textit{Integrated Gradients} (IG) \cite{DBLP:conf/icml/SundararajanTY17}. Like Shapley values, the method satisfies the conservation property $\sum_i R_i = f(\x)$. An advantage of the integrated gradients approach is that it does not require evaluating the function exponentially many times, and each gradient evaluation readily provides a $d$-dimensional explanatory feedback without having to test each input feature separately. The number of required function evaluations corresponds to the level of discretization of the integral. However, in comparison to the Shapley value approach, IG only explores a small region of the input space, typically the segment connecting the reference point to the data point. Hence, the explanation may be biased by this local scope and it may therefore fail to integrate important components of the decision process.

Approaches such as SmoothGrad \cite{DBLP:journals/corr/SmilkovTKVW17} mitigate the problem by repeating the analysis multiple times with some randomness factor (in the case of IG, one can for example randomize the integration path \cite{Erion2021, DBLP:journals/pieee/SamekMLAM21}). This comes however at an increased computational cost. For a review of gradient-based explanation techniques, we refer the reader to \cite{DBLP:series/lncs/AnconaCOG19}.

\subsection{Propagation-based explanations}
\label{section:methods-propagation}

A last category of methods aims to leverage the neural network structure to produce explanations \cite{DBLP:conf/eccv/ZeilerF14, DBLP:journals/corr/SpringenbergDBR14, bach-plos15, DBLP:conf/icml/ShrikumarGK17}. The layer-wise relevance propagation (LRP) method \cite{bach-plos15}, in particular, solves the explanation task by starting at the output of the network, and reverse propagating the prediction, layer by layer, until the input variables are reached. The propagation at each layer is implemented by a purposely designed propagation rule. Let $j$ and $k$ denote indices of neurons at two consecutive layers, and let $a_j$ and $a_k$ denote their respective activations, with $w_{jk}$ the weight connecting these two activations. Denote by $R_k$ the `relevance' received by neuron $k$ from the layers above, which can be interpreted by the contribution of neuron $k$ in its corresponding layer to the output prediction $f(\x)$. The propagation rules used by LRP are typically of the form:
$$
R_j = \sum_k \frac{z_{jk}}{\sum_{0,j} z_{jk}} R_k
$$
where $z_{jk}$ models the contribution of neuron $j$ to the activation of neuron $k$, where $\sum_k$ is a sum over neurons of the current layer, and where $\sum_{0,j}$ is a sum over neurons in the layer below plus the bias represented as an additional neuron with activation $a_{0} = 1$. It is easy to show that when the neural network does not have biases (or if biases are not included in the propagation rule), we have conservation from layer to layer, i.e.\ we can build the chain of equalities
$$
\textstyle \sum_i R_i = \dots = \sum_j R_j = \sum_k R_k = \dots = f(\x)
$$
which results in the same conservation property as for the Shapley value or integrated gradients.

\smallskip

In practice, various LRP rules can be used. For deep networks with ReLU activations, LRP-$\gamma$ \cite{DBLP:series/lncs/MontavonBLSM19} was proposed, and sets $z_{jk} = a_j \cdot (w_{jk} + \gamma w_{jk}^+)$. The hyperparameter $\gamma$ can be tuned to improve robustness of the explanation. Choosing $\gamma=0$ yields typically noisy explanations that coincide with a simple `Gradient$\,\times\,$Input' explanation. Choosing $\gamma$ larger, especially in lower layers, yields more robust explanations that become empirically closer to those of the Shapley value.---Here, the main advantage of LRP is that explanations can be produced in a single forward\,/\,backward pass, which makes the method applicable to highly complex models with hundreds or thousands of input features, such as convolutional neural networks (CNNs) used in vision.

\smallskip

The LRP method, although most studied on deep ReLU networks, can also be applied or extended to different types of models, e.g.\ long short-term memory networks (LSTMs) \cite{DBLP:series/lncs/ArrasAWMGMHS19}, graph neural networks (GNNs) \cite{schnake2020xai}, Bayesian Neural networks \cite{bykov2020much} or non-neural networks, e.g.\ for anomaly detection \cite{DBLP:journals/pr/KauffmannMM20,DBLP:journals/pieee/RuffKVMSKDM21} or clustering \cite{DBLP:journals/corr/abs-1906-07633}. For a detailed overview of LRP, we refer the reader to \cite{DBLP:series/lncs/MontavonBLSM19,DBLP:journals/pieee/SamekMLAM21}.

\bigskip

\begin{figure*}[t!]
\centering
\includegraphics[width=\textwidth]{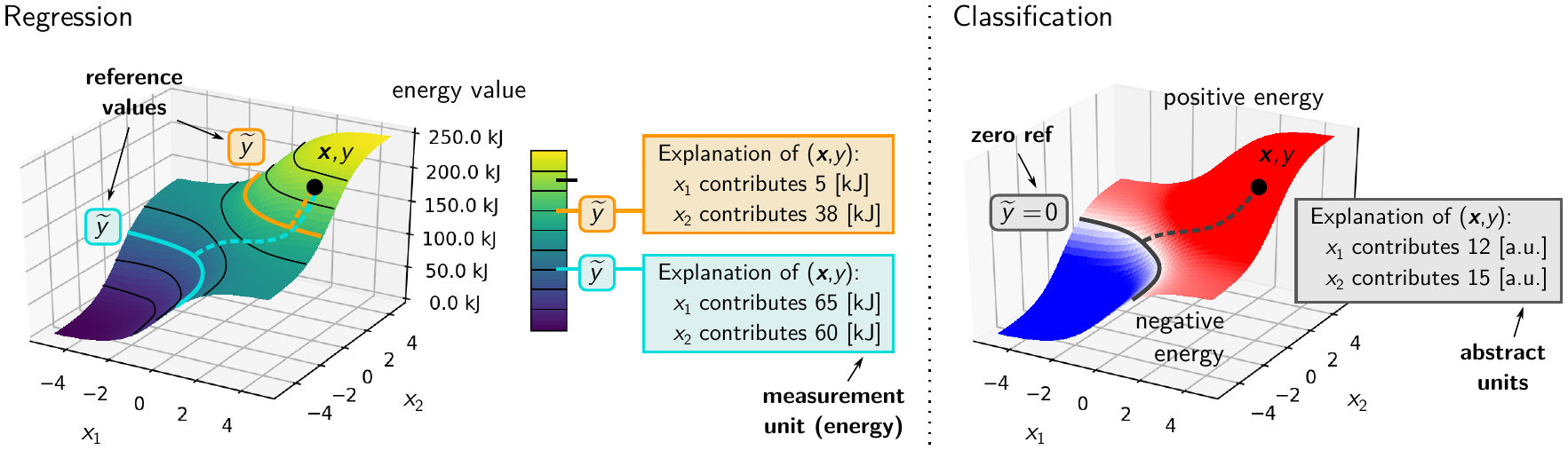}
\caption{Overview of Explainable AI for regression and for classification. The variable $\x$ denotes some point of interest for which we would like to explain the prediction $y = f(\x)$. XAI for regression differs in several ways from the usual classification scenario: (1) the explanation scores can be interpreted as physical quantities of the same units as the prediction; (2) the explanation is produced relative to some user-defined reference value $\widetilde{y}$, and the latter can substantially affect the explanation.}
\label{fig:overview}
\end{figure*}

\subsection{Other methods}

For comprehensiveness, we also briefly mention some other approaches to explanation. Some explanations are defined as the result of an optimization problem (e.g.\ \cite{DBLP:conf/iccv/FongV17}). Certain explanation methods address the slightly different task of extracting a subset of relevant input features rather than scoring these features \cite{DBLP:conf/icml/ChenSWJ18,DBLP:conf/iclr/YoonJS19}. Some explanations are obtained by training a local surrogate model which is easier to explain, e.g.\ a linear model or a decision tree (e.g.\ \cite{DBLP:conf/kdd/Ribeiro0G16}). Some explanation methods assume a particular self-explainable structure (e.g.\ additivity or attention mechanisms) in the model \cite{caruana15,DBLP:conf/cvpr/ZhouKLOT16,DBLP:conf/iclr/BrendelB19,Rudin2019,xu2015show}. Finally, some methods, commonly referred as higher-order methods aim to identify not individual features, but groups of features, that contribute only when occurring jointly \cite{grabisch1999axiomatic,DBLP:conf/iclr/TsangC018,Eberle-BiLRP,schnake2020xai,DBLP:journals/corr/abs-1901-08361,lundberg2020local}.
Moreover, methods have been proposed or adjusted to fit specific model architectures (e.g. LSTMs \cite{DBLP:series/lncs/ArrasAWMGMHS19} or GNNs \cite{DBLP:conf/nips/YingBYZL19,schnake2020xai}) as well as characteristics connected to different data types, such as time-series \cite{Cho21,lim2021temporal, DBLP:journals/corr/abs-2104-00950} or natural language \cite{danilevsky2020survey}.

In parallel to the broadening of explanation methods, there have been several works providing unifying views on the different explanation techniques \cite{DBLP:conf/nips/LundbergL17, DBLP:series/lncs/AnconaCOG19,DBLP:journals/pieee/SamekMLAM21,Covert2021}, drawing connections between gradient-based and perturbation-based explanation methods, as well as showing how propagation-based methods reduce to gradient-based methods for certain choices of parameters. 

\section{XAI for Regression (XAIR)}
\label{section:motivating}

Regression is an important subfield of ML and signal processing, which focuses on predicting quantities that are real-valued instead of categorical. Such real-valued predictions are predominant in many practical application scenarios, often related to physics, engineering or economics. Explainable AI techniques have only recently started to be applied in such regression scenarios. For example, \cite{Jahromi_InvTemp_2020} have applied the Shapley value framework to shed light on a heat transfer phenomena modeled by a ML model. In the area of hydrology, \cite{DBLP:series/lncs/KratzertHKHK19} are making use of integrated gradients to identify what factors (precipitation, temperature, radiation, etc.) contribute to river discharge, as modeled by an LSTM neural network. In the area of energy engineering, \cite{Papadopoulos_BuildEn_2019} use Shapley values to attribute building energy consumption, as predicted by the XGBoost ML model, to input features such as unit density, number of floors, or built year. Finally, \cite{schnake2020xai} contributes an extension of the LRP method to GNNs, and applies it in a learning scenario for quantum chemistry to explain molecular atomization energy in terms of individual atoms or groups of atoms.

\medskip

So far, the most common way of applying XAI to regression has been to use solutions developed in the context of classification, and apply them to the regression case, either by using them in an out-of-the-box manner (e.g.\ \cite{Jahromi_InvTemp_2020,Papadopoulos_BuildEn_2019,DBLP:series/lncs/KratzertHKHK19,schnake2020xai}) or by translating a regression into a multi-class classification problem (e.g.\ \cite{lapuschkin2017understanding, binder2021morphological}). While this direct approach is technically straightforward and has delivered useful practical insights, more precise and understandable explanations can be gained from carefully revisiting Explainable AI in the specific context of regression, and some initial steps have been taken in the particular area of counterfactual explanations \cite{spooner2021counterfactual, hadaexploring}.

\medskip

In this paper, we address the problem of XAI for regression in a broad manner and identify two important specificities of the regression problem that require an adaptation of XAI: The first specificity has to do with the nature of the prediction itself, where the prediction of a regression model is often attached with a particular \emph{measurement unit} (e.g.\ physical or monetary). If using an appropriate explanation method, the unit of the prediction can be inherited by the explanation, which enables further interpretability of the explanation result for the user. The second specificity relates to the typically higher amount of information contained in a real-valued prediction compared to a classification (especially when the regression task has a high signal-to-noise ratio). For the explanation to faithfully address the user's need, further contextualization is required, for example, by specifying a particular \emph{reference value}. These two aspects are illustrated in Fig.\ \ref{fig:overview}. We elaborate on them in more detail in the sections below.

\subsection{Explaining Quantities with Units (e.g.\ Physical\,/\,Monetary)}

The first aspect that distinguishes the task of explaining an ML regression model from that of explaining an ML classifier is the nature of the prediction itself, and what can we gain from attributing the prediction to the input features.

In the classification case, the output of the ML model can take various forms, for example, a decision outcome, a logit score, a class probability, a distance to the decision boundary, or else. For most of them, these forms are rather abstract and hard to interpret for the end-user. This is also the case for an attribution of these quantities to the input features, and consequently, these attributions are often used merely as a visualization (e.g.\ a histogram or a heatmap). Alternatively, simpler forms of explanation such as a list of the top-k most relevant features or some visual mask in pixel space, have also shown to be efficient for explaining classification outcomes. This preference for simple explanations is also reflected in the design of XAI benchmarks, where only the set of most relevant features or the ordering of features from most to least relevant usually enters into consideration \cite{SamTNNLS17,DBLP:conf/iclr/YoonJS19,DBLP:conf/nips/YingBYZL19}.

In the regression case, however, where we typically predict real-valued quantities such as price, energy, etc., one has the opportunity to retain the measurement unit in the explanation. In a real-world system, information such as the value (e.g.\ in some monetary unit) attributed to a particular subentity, or the amount of energy attributed to a particular interaction or a subsystem, have a more direct use. They may provide a mechanism for pricing, or for predicting what amount of energy might be transferred from one subsystem to another, e.g.\ a chemical reaction. To ensure this physical interpretation of attributions, the property of conservation (i.e.\ $\sum_i R_i = f(\x)$) becomes crucial.\footnote{Note, that with a simple scaling of attributions any explanation method can formally fulfill the conservation criterion, yet, it is unlikely that the rescaled scores truly reflect the contribution of each variable to the predicted score.} Decomposition methods such as the Shapley value, integrated gradients or LRP, which we have introduced in Section \ref{section:review}, are then strongly preferred over methods such as sensitivity analysis or feature selection, which per se do not fulfill this property. 

\subsection{Explaining with a User-Provided Reference Value}\label{sec_ref_value}

The second aspect that separates classification and regression from the perspective of explanation, has to do with the type of output domain, in particular, whether it is categorical or real-valued.

In classification, we often like to formulate the problem of explanation in terms of a baseline scenario, e.g. ``what makes a deep neural network predict there is a cat in an image vs.\ not a cat''. Such questions can be addressed in technical terms by considering the logit (or probability) score $f(\x)$ at the output of the network, and analyze it relative to some \textit{fixed} value, the decision threshold, representing where the decision ``cat'' transitions to ``not a cat''. This is roughly the regime at which techniques such as the Shapley value, integrated gradients, and LRP usually operate.

In regression, however, outputs are not categorical and the question arises naturally with what exactly we should contrast the prediction of interest. In this paper, we argue that a \textit{reference value} should be specified as part of the question for which the user seeks an explanation. For example, the user may ask ``why a physical system is predicted to have an energy of \SI{2500}{\kilo\joule/\mol} vs. \SI{1000}{\kilo\joule/\mol}'', or ``why an item is currently valued at 1200 dollars compared to its usual 1000 dollars price''. XAIR methods thereby enable the user to get a contextualized explanation with respect to a specific reference value. This delivers more targeted explanations than obtained by existing XAI methods with their often implicit and fixed reference values. We will further motivate and illustrate the benefits of extending XAI frameworks in this manner in the following section.

\subsection{Motivating Example: Auction Scenario}
\label{sec:motivating_example}
\begin{figure}[t]
    \centering
    \includegraphics[width=\linewidth]{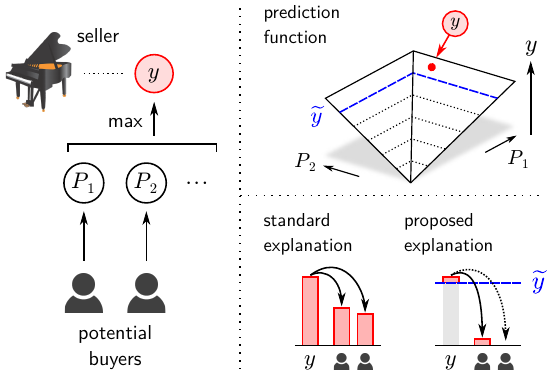}
    \caption{Auction scenario. The ML function predicts the outcome (selling price) of the auction. The prediction must be explained in a meaningful manner in terms of the individual buyers and their features.}
    \label{fig:auction}
\end{figure}

To illustrate the specificity of explanation in the regression context, we consider a toy example consisting of a first-price auction scenario, where a seller would like to sell some item at the highest possible price to the auction participants (see also Fig.\ \ref{fig:auction}). Let $P_1,\dots,P_d$ describe the prices (expressed in some given monetary unit) at which the $d$ participants place their bids. We assume a typical regression problem where these prices are predictable using some function $P_i = \mathcal{P}(\x_i)$ where $\x_i$ is a representation of the $i$th participant, e.g.\ based on demographic features or bidding history.

The outcome of the first-price auction is given by the maximum value between the $d$ bidders, i.e.
$$
y = \max(P_1,\dots,P_d).
$$
For simplicity we assume two bidders ready to purchase at price $P_1 = 1100$ and $P_2 = 900$ monetary units, respectively. Note that if we would produce an explanation with a reference price of $\widetilde{y} = 0$, both bidders would be attributed roughly half of the money as they share responsibility in setting the price higher than zero.

However, if we assume that the seller sets a base price of $1000$ monetary units (e.g.\ corresponding to the value of the item according to the seller), and that the seller does not sell the item if no one bids above that base price, the question to answer becomes how to explain the benefit of the seller, i.e.\ ``why the selling price is $y$ and not $1000$''. In other words, we want to know what is the explanation of the model output when considering a reference value $\widetilde{y} = 1000$. Analyzing the real-valued function $P$ at $\widetilde{y}=1000$ intuitively reveals that only the first bidder contributes to the benefit of the seller as the second bidder is unwilling to buy the item even at the base price. (We will present methods in Section \ref{sec:method} to produce these explanations systematically.)

An alternative approach would have been to explain the classification decision $P>1000$ based on standard XAI techniques. However this approach is limited by the fact that multiple functions (e.g.\ $P$, $\tanh(P-1000)+1000$, $P^2/1000$, etc.) support the same decision boundary. Yet most of them do not correctly reflect the price away from the decision boundary. Using them would lead to distorted explanations, that would not reliably identify buyers' contributions to the selling price. This problem is fully avoided by explaining the regression function directly.

Overall, with the help of a regression-based explanation, the seller acquires a precise knowledge of the amount each participant of the auction has contributed to the selling price, specifically, to the difference between the selling price and the base price (or the seller's personal benefit). These scores can then be further attributed to e.g.\ the participants' demographic features, so that one can identify in which demographic group future auctions should be advertised for maximum gain.

\section{Bringing XAI Methods to Regression}
\label{sec:method}

The motivating example above has highlighted how important the choice of reference value $\widetilde{y}$ and the conservation of units are in the regression case. They are an integral part of formulating the question the user seeks an explanation for. State-of-the-art explanation techniques, however, still lack the possibility to incorporate user-provided reference values $\widetilde{y}$ and therefore the flexibility to meaningfully incorporate this added information.

Our starting point is to state the function we wish to explain. We call such function $g(\x)$, and define it as:
\begin{align}
g(\x) = f(\x) - \widetilde{y}
\label{eq:centering}
\end{align}
i.e.\ the original function $f(\x)$ \textit{relative} to the user-specified reference value $\widetilde{y}$. We now consider whether existing XAI techniques, in particular, techniques presented in Section \ref{section:review}, readily apply to this newly defined function, or whether modifications of these techniques are needed.

\begin{figure}[h]
    \centering
    \includegraphics[width=\linewidth]{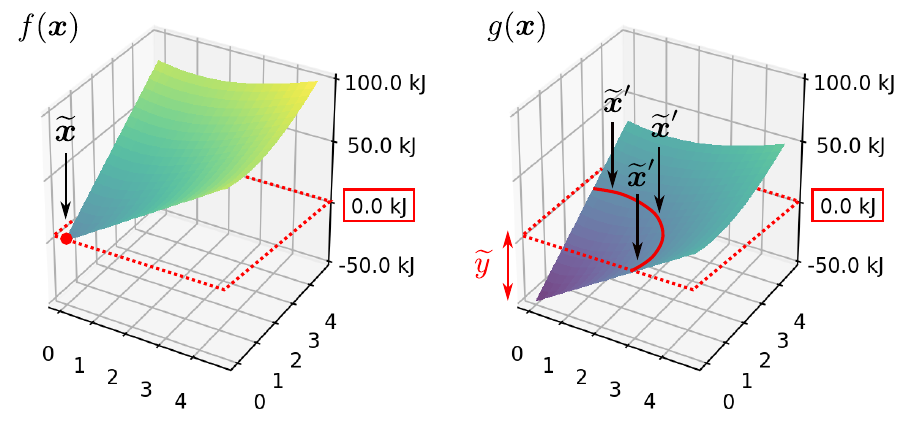}
    \caption{Effect of shifting the function on the function's root points. After shifting, the original root point $\widetilde{\x}$ is no longer a root point, and there are potentially many new root points to choose from.}
    \label{fig:centering}
\end{figure}

\textit{Removal-based} and \textit{gradient-based} explanation methods (presented in Sections \ref{section:methods-removal} and \ref{section:methods-gradient}) are designed in a way that they can apply to \textit{any} function. These methods typically only require a root point $\widetilde{\x}$ at which the function has value zero. We note, however, that if $\widetilde{\x}$ is a root point of the function $f(\x)$, it is typically \textit{not} a root point of the function $g(\x)$, and the function $g(\x)$ has potentially multiple new root points $\widetilde{\x}'$ to choose from. Such dilemma is highlighted in Fig.\ \ref{fig:centering}.

 Choosing one particular root point $\widetilde{\x}'$ might introduce some spurious modeling bias into the explanation. Alternately, one can consider a set of root points (e.g.\ weighted according to some probability distribution) and compute the expectation over the resulting explanations. This reduces (although not fully eliminates) the risk of modeling bias but results in a significant increase of computational cost. Lastly, one can tamper with the function $g(\x)$. For example, when $0 \leq \widetilde{y} \leq f(\x)$, one can apply the clipping nonlinearity $g^+(\x) = \max(0,g(\x))$ to ensure that the original root point remains a root point, i.e.\ $g^+(\widetilde{\x}) = 0$. Consequently, one can use $\widetilde{\x}$ for the analysis.\footnote{A similar construction can be obtained for the case $f(\x) \leq \widetilde{y} \leq 0$, where we define instead the function $g^-(\x) = \min(0,g(\x))$.} While this avoids introducing explanation bias through the choice of a new reference point, the rectification function itself may introduce bias, although less significant. Furthermore, this tampering approach is only applicable when the value $\widetilde{y}$ is located between $0$ and $f(\x)$.

In contrast to \textit{removal-based} or \textit{gradient-based} explanation methods, \textit{propagation-based} methods such as LRP (cf.\ Section \ref{section:methods-propagation}) do not require a reference point $\widetilde{\x}$ to produce an explanation. Propagation-based methods rely instead on the neural network's internal representation of the function to progressively redistribute the prediction from layer to layer until the input features are reached. These methods assume, however, that the neural network supporting the explanation is reasonably disentangled. While the newly defined function $g(\x)$ can be seen as a neural network (by simply adjusting the top-layer bias), the neural network representation may not be sufficiently disentangled to explain subtle variations of the function $f(\x)$ around the provided reference value $\widetilde{y}$. Furthermore, in Explainable AI, biases are typically having the role of `unexplained' factors \cite{DBLP:series/lncs/AnconaCOG19}, and thus, the explanation of such bias-adjusted model would essentially remain the same except for an unexplained term.

We propose two strategies to enable propagation-based methods to explain $g(\x)$: First, \textit{retraining}, where a surrogate neural network is trained to replicate $g(\x)$ on some relevant band of prediction values. Second, \textit{restructuring}, where the last few layers are rewritten (without retraining) in a way that the network outputs $g(\x)$, and the representation has been sufficiently altered to enable the desired fine-grained explanation. These two approaches are presented in detail below.

\subsection{Retraining}
\label{sec:retraining}

The first approach consists of retraining the network, so that it accurately predicts the new function $g(\x)$ on some relevant band of function values $\tau^- \leq y \leq \tau^+$. In particular, we can define some surrogate neural network model $\hat{g}(\x,\theta)$, where $\theta$ represents the set of parameters of the network, and optimize the objective
$$
\mathcal{E}(\theta) =  \sum_{n=1}^N \big(\hat{g}(\x_n;\theta)-g(\x_n)\big)^2 \cdot 1_{\{\tau^- \leq g(\x_n) \leq \tau^+\}}
$$
where $1_{\{\cdot\}}$ is an indicator function. To prevent the learning algorithm to only recalibrate top-layer biases, we propose to freeze these biases to their original value, or to incorporate additional penalties to the objective function.

Note that unlike a classification-based retraining approach, where the model would be instructed to classify each example into bins representing different ranges of regression values, the approach described here preserves the unit of measurement (e.g.\ price or energy) at the output of the original model. This also holds for explanations of such regression models, where attribution on the input features produces scores expressible in the same unit.

For the retraining approach to deliver good results, one needs to pay particular attention to the data used for retraining. Choosing a dataset that is too limited may result in a model that predicts in the same way but is prone to the apparition of Clever Hans effects \cite{LapNCOMM19,anders2022finding} which may unfaithfully explain the original prediction function. Conversely, the new model may also inhibit preexisting Clever Hans effects from the original model concealing its weaknesses and preventing a meaningful model validation. Lastly, for complex models that require large training sets, retraining results in a high computational cost, especially if a large number of reference values are of interest. 

\subsection{Restructuring}
\label{sec:restructuring}

To avoid the retraining step, an alternative approach consists of manually restructuring the network in a way that the network output becomes exactly $g(\x)$. Restructuring should be profound enough for the neural network internal representation to be meaningfully affected and capable of influencing the explanation process in the desired manner. The restructuring approach is remotely related to `neuralization' \cite{DBLP:journals/pr/KauffmannMM20,DBLP:journals/corr/abs-1906-07633,DBLP:journals/pieee/RuffKVMSKDM21}, which converts non-neural network models (e.g.\ kernel machines) into functionally equivalent neural network models, without retraining.

Our restructuring approach assumes that the last two layers of the neural network are ReLU and linear respectively. (This includes the special case where the linear layer is an average pooling layer or a combination of a linear and an average-pooling layer.) Restructuring is implemented as a backward pass, where the reference value adjustment of the function $g(\x)$ is propagated first in the linear layer, and then in the ReLU layer. These two steps of propagation, which we describe below, are profound enough to meaningfully change the representation and the produced explanation.

\smallskip

\subsubsection*{Step 1: Propagation in the linear layer} Denote by $y = \sum_j a_j w_j + b$ the equation for the original top-level linear layer. The shifted top-layer can be rewritten as the same linear layer with the reference value $\widetilde{y}$ transformed into a reference activation vector $\widetilde{\ba}$:
\begin{align}
y-\widetilde{y} &=\textstyle \sum_j a_j w_j + b -\widetilde{y} \nonumber\\
&= \textstyle \sum_j (a_j - \widetilde{a}_j) \cdot w_j + b
\label{eq:prop-taylor}
\end{align}
For this equation to hold, we need to choose $\widetilde{\ba}$ such that $\sum_j \widetilde{a}_j w_j = \widetilde{y}$. In practice, there are many possible choices for $\widetilde{\ba}$.
We propose a `flooding' strategy where the reference point $\widetilde{\ba}$ is chosen in a way that only large neuron activations (corresponding to global effects) remain in $\widetilde{\ba}$. Specifically, we search for a reference point $\widetilde{\ba}^{\text{(flood)}}$ at the intersection of Eq.\ \eqref{eq:prop-taylor} and the parameterized line:
$$
\{(\ba - t \cdot \boldsymbol{1})^+,t \in \mathbb{R}\}
$$
where $(\cdot)^+$ applies element-wise. This reference point absorbs all large activations (coarse-grained effects), so that the explanation is contextualized and can focus on small activations representing fine local variations.

\medskip

\begin{example*}
To illustrate the effect of the restructuring approach on the internal representation and subsequently on the explanation, we consider the auction example of Section \ref{sec:motivating_example}. In the case where there are two buyers, the max function between two positive values can be written as a neural network using the composition of ReLU neurons:
$$
f(\x) = \frac12 {\underbrace{(x_1+x_2)^+}_{a_3}} + \frac12 {\underbrace{(x_1-x_2)^+}_{a_4}} + \frac12 {\underbrace{(x_2-x_1)^+}_{a_5}}
$$
In the case where the two actors bid a similar price, e.g. $1100$ and $900$ monetary units respectively, the neuron activation $a_3$ modeling the coarse effect (i.e.\ the price average) has a much higher value than $a_4$ and $a_5$ that model the fine effect (i.e.\ the price difference). This implies that the application of our restructuring approach to this example will significantly reduce the influence on $a_3$ on the explanation but preserve the influence of $a_4$ and $a_5$. This modified representation helps in turn to deliver the desired explanation where the difference between the two bidders, modeled by the neuron $a_4$, is now more strongly expressed.
\end{example*}

Note that the proposed restructuring approach is only directly applicable to a network with a single output because the reference point $\widetilde{\ba}$ depends on the weights of the output neuron. In presence of a multi-output neural network, different restructurings need to be applied for each output neuron.

\medskip

\subsubsection*{Step 2: Propagation in the ReLU layer} 
\begin{figure}[h]
    \centering
    \includegraphics[width=\linewidth]{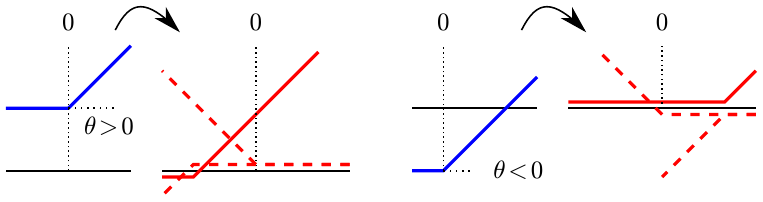}
    \caption{Transformation of shifted ReLUs neuron (blue) into a sum of multiple non-shifted ReLUs (red). The latter is functionally equivalent but more amenable to the LRP propagation technique.}
    \label{fig:relu}
\end{figure}

The restructuring step we have performed above implies that each ReLU neuron in the layer below is now attached with an offset $\theta= - \widetilde{a}_j$. Such a shifted ReLU neuron is not directly explainable within common propagation-based explanation frameworks such as LRP. We observe however that we can rewrite the shifted ReLU activation as a linear combination of three standard ReLU activations:
\begin{align*}
\rho(z_j) - \widetilde{a}_j &=\rho(z_j - \widetilde{a}_j) + \rho(-z_j) - \rho(-z_j + \widetilde{a}_j)
\end{align*}
where $\rho$ denotes the ReLU nonlinearity. The ReLU neuron and its restructuring are depicted in Fig.\ \ref{fig:relu}. In practice, each neuron in the given layer is first triplicated; then all outgoing weights of that layer are multiplied by $1$, $1$, and $-1$ for each neuron copy respectively; incoming weights are multiplied by $1$, $-1$ and $-1$; and finally, the offset $\widetilde{a}_j$ is multiplied by $-1$, $0$, and $1$ before being included in the corresponding neuron biases.

With these two steps of restructuring, we arrive at a functionally equivalent architecture composed exactly of the same layer types, and where the number of neurons in the last hidden layer has increased by a factor of three. The main advantage of the resulting neural network is that it supports a better redistribution of the output score on these layers (e.g., using LRP). Furthermore, the restructuring approach guarantees that the function to explain remains the same.

Compared to the retraining approach, the restructuring approach avoids an inadvertent modification of the model's decision strategy through a suboptimal retraining algorithm or an improper choice of data distribution for retraining. As a downside, the restructuring approach requires a particular top-layer structure, which makes the approach less model-independent. In practice, however, many state-of-the-art models, for example, neural networks for computer vision have this top-layer structure.

\section{Validation Experiments}
\label{section:evaluation}

In this section, we will evaluate the two proposed XAIR approaches, retraining and restructuring, in combination with the LRP explanation technique. Produced explanations will be evaluated on a collection of low-dimensional datasets where the application of the Shapley value remains computationally feasible and can (for these scenarios) serve as a reference explanation method for comparison. Our two proposed approaches will be compared against two baselines which simply consist of shifting or scaling the original explanation. We will demonstrate that our two approaches perform substantially better than the baselines. In particular, our restructuring approach does not incur any retraining cost and thus runs as quickly as the original LRP procedure.

\subsection{Datasets}
For our validation experiments, we make use of 5 different low-dimensional datasets (between 4 and 13 dimensions), referred to as \textit{max}, \textit{linear}, \textit{friedman}, \textit{diabetes}, and \textit{boston}. The \emph{max} dataset corresponds to our motivational example of Section \ref{sec:motivating_example} where we compute $y=\max(x_1,\dots,x_d)$ from $d=8$ uniformly distirbuted features. The \emph{linear}\footnote{\url{https://scikit-learn.org/stable/modules/generated/sklearn.datasets.make_regression.html}} dataset is a simple linear regression problem where 4 out of 8 input dimensions have predictive power. The \emph{friedman}\footnote{\url{https://scikit-learn.org/stable/modules/generated/sklearn.datasets.make_friedman2.html}} regression problem is another synthetic dataset where a non-linear function is predicted from independent uniformly distributed features \cite{FiedmannDataset91}. The \emph{diabetes}\footnote{\url{https://scikit-learn.org/stable/modules/generated/sklearn.datasets.load_diabetes.html}} dataset is a well-known regression example composed of 442 instances, where one has to predict disease progression from $10$ patient features. Finally, the \emph{boston housing}\footnote{\url{https://scikit-learn.org/stable/modules/generated/sklearn.datasets.load_boston.html}} dataset consists of 506 instances of housing values in Boston, which are to be predicted from 13 geographical features. To facilitate optimization (and without any loss of generality), for all experiments, we standardize all features independently and rescale targets such that they are between zero and one (the learned model can be rescaled back to the original units after training for the purpose of explanation).

\subsection{Experimental setup}
We train on each dataset a two-layer neural network, consisting of an intermediate layer of 256 ReLU neurons and one linear neuron as output. We optimize weights to minimize the mean square error between prediction and target, using stochastic gradient descent (varying number of epochs and learning rates for different datasets). Overall, our trained models are able to solve the tasks reasonably well with coefficients of determination (R$^2$) higher than 0.9.

Given the simple low-dimensional datasets used in our experiment, we can generate reference explanations by applying the Shapley value \cite{Shapley1953,DBLP:journals/jmlr/StrumbeljK10}, which is expensive to compute but theoretically well founded. Because the Shapley value is a removal-based explanation technique, we adopt the clipping strategy $g^+(\x)$, one of the options proposed in Section \ref{sec:method}), which introduces the minimum amount of bias into the explanation. Consequently, we only retain in our evaluation instances where the clipping approach is applicable, i.e.\ satisfying $0 \leq \widetilde{y} \leq f(\x)$. We then generate attributions for the respective reference values using our two proposed methods (retraining and restructuring). For the retraining approach, we set the parameter controlling the band of function values to $\tau^-=-0.3$ and $\tau^+=\infty$, and we initialize the layers with the weights from the original model. In both cases we use the LRP-$\gamma$ rule with $\gamma=2.5$ and $\gamma=0$ for the first and second layers. We compare the proposed methods against two simple baselines that represent a simple post-processing of a standard LRP explanation (without training or restructuring), in particular, our `shift' baseline computes the new relevance scores $R_i' = R_i - \widetilde{y}/d$ and our `scaling' baseline computes $R_i' = R_i \cdot (y-\widetilde{y})/y$. We evaluate our method for different reference values $\widetilde{y}_q = q f(\x)_{\max} + (1-q) f(\boldsymbol{0})$.

Explanation performance is measured as the mean square error (MSE) between the produced explanation and the Shapley-based reference explanation. Unlike many explanation evaluation metrics (e.g.\ based on IOU scores \cite{DBLP:journals/ijcv/ZhangBLBSS18} or pixel-flipping curves \cite{SamTNNLS17}), the MSE explicitly accounts for the magnitude of the feature attributions and not only their ordering. When reporting the results, we also divide the MSE score by the average MSE of random attributions scaled to the difference between $y_i$ and $\widetilde{y}$. A value of 1, therefore, represents the error equivalent to a random assignment of attributions that satisfy the completeness property. Error bars are calculated by repeating every experiment 10 times.

\begin{table*}[h!]
    \centering
\caption{
Results for validation experiments. Performance measured in normalized MSE. Inputs normalized, targets scaled between 0/1, and model with 256 neurons in hidden layer with ReLU-activations.}
    \begin{tabular}{ll|rrrr}
\toprule
& & \multicolumn{2}{c}{Baselines} & \multicolumn{2}{c}{Our methods}\\
    dataset &    &                  explanation shift &                 explanation scaling &   model retraining &  model restructuring \\
\midrule
    max & $\widetilde{y}_{0.25}$ &    1.0343 $\pm$ 0.171 &  0.3316 $\pm$ 0.015 &  \textbf{0.2763 $\pm$ 0.014} &           0.2968 $\pm$ 0.016 \\
        & $\widetilde{y}_{0.5}$  &    4.9424 $\pm$ 0.703 &  0.2603 $\pm$ 0.017 &  \textbf{0.2042 $\pm$ 0.015} &           0.2102 $\pm$ 0.017 \\
        & $\widetilde{y}_{0.75}$  &   22.9945 $\pm$ 3.097 &  0.2380 $\pm$ 0.022 &           0.2022 $\pm$ 0.016 &  \textbf{0.1979 $\pm$ 0.020} \\\midrule
    linear & $\widetilde{y}_{0.25}$ &    2.3851 $\pm$ 0.411 &  0.4008 $\pm$ 0.030 &  \textbf{0.3179 $\pm$ 0.026} &           0.3549 $\pm$ 0.040 \\
        & $\widetilde{y}_{0.5}$ &    8.5959 $\pm$ 1.967 &  0.3407 $\pm$ 0.049 &  \textbf{0.2450 $\pm$ 0.037} &           0.2674 $\pm$ 0.054 \\
        & $\widetilde{y}_{0.75}$ &  57.8571 $\pm$ 47.084 &  0.4731 $\pm$ 0.078 &           0.4208 $\pm$ 0.139 &  \textbf{0.3747 $\pm$ 0.042} \\\midrule
    friedman & $\widetilde{y}_{0.25}$ &    0.8433 $\pm$ 0.109 &  0.2067 $\pm$ 0.037 &           0.1835 $\pm$ 0.038 &  \textbf{0.1345 $\pm$ 0.047} \\
        & $\widetilde{y}_{0.5}$ &    3.5789 $\pm$ 0.788 &  0.2194 $\pm$ 0.044 &           0.1525 $\pm$ 0.042 &  \textbf{0.1241 $\pm$ 0.040} \\
        & $\widetilde{y}_{0.75}$ &  47.0557 $\pm$ 16.629 &  0.2714 $\pm$ 0.121 &  \textbf{0.1723 $\pm$ 0.105} &           0.2017 $\pm$ 0.052 \\\midrule
     diabetes & $\widetilde{y}_{0.25}$ &    0.6823 $\pm$ 0.080 &  0.3547 $\pm$ 0.037 &           0.3205 $\pm$ 0.040 &  \textbf{0.3122 $\pm$ 0.024} \\
        & $\widetilde{y}_{0.5}$ &    2.8446 $\pm$ 0.524 &  0.3443 $\pm$ 0.082 &           0.2835 $\pm$ 0.075 &  \textbf{0.2784 $\pm$ 0.048} \\
        & $\widetilde{y}_{0.75}$ &    7.0100 $\pm$ 2.107 &  0.3075 $\pm$ 0.126 &  \textbf{0.2383 $\pm$ 0.102} &           0.2489 $\pm$ 0.068 \\\midrule
     boston & $\widetilde{y}_{0.25}$ &    0.7655 $\pm$ 0.111 &  0.3593 $\pm$ 0.029 &           0.3553 $\pm$ 0.043 &  \textbf{0.2779 $\pm$ 0.032} \\
        & $\widetilde{y}_{0.5}$ &    2.9533 $\pm$ 0.384 &  0.3819 $\pm$ 0.041 &           0.3315 $\pm$ 0.044 &  \textbf{0.2662 $\pm$ 0.048} \\
        & $\widetilde{y}_{0.75}$ &  43.8898 $\pm$ 15.818 &  0.8289 $\pm$ 0.098 &           0.8238 $\pm$ 0.123 &  \textbf{0.6149 $\pm$ 0.133} \\
\bottomrule
\end{tabular}
    \label{tab:rmse_eval}
\end{table*}

\subsection{Results}
\begin{figure}[h]
    \centering
    \includegraphics[width=\linewidth]{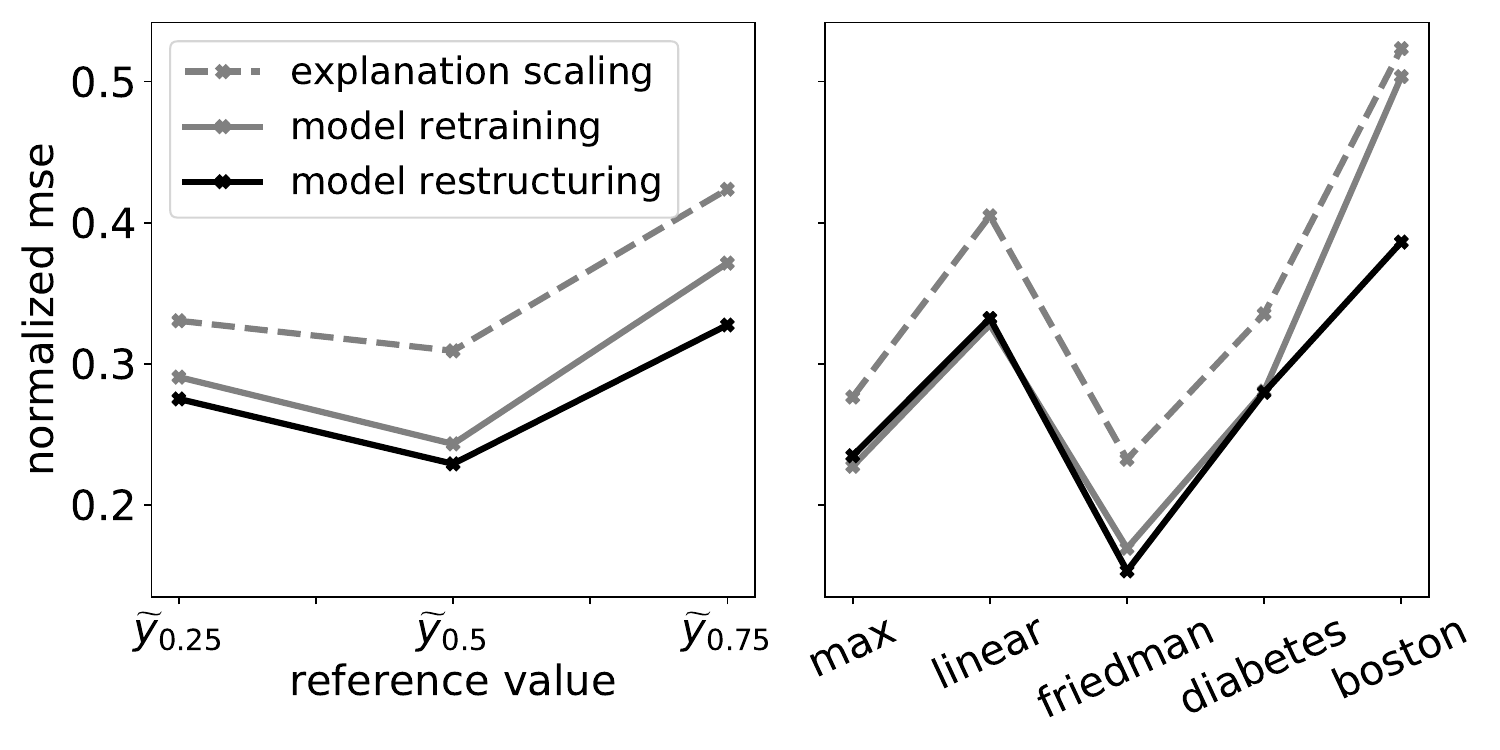}
    \caption{Mean normalized explanation error (in MSE) of methods by choice of reference value $\widetilde{y}_q$ (left) and dataset (right).}
    \label{fig:eval_scalevsflood}
\end{figure}

Table \ref{tab:rmse_eval} evaluates the proposed methods and baselines for the different datasets over a selection of reference values. It can be observed that the proposed restructuring and retraining approaches outperform the baselines for all configurations. Figure \ref{fig:eval_scalevsflood} (left) shows the average normalized MSE per $\widetilde{y}_q$, excluding the shift-baseline due to its poor performance. Figure \ref{fig:eval_scalevsflood} (right) shows the average normalized MSE per dataset, again without the shift-baseline. Average performance is qualitatively consistent across all datasets and reference values with model restructuring in the lead. In absolute terms, all methods show a higher MSE on the boston dataset compared to the others.

When comparing `model restructuring' to `explanation scaling' (the closest competitor among baselines) we observe an overall decrease in MSE of more than 22\%. The advantage ranges from around 15\% for the max dataset up to 34\% for the friedman dataset. In terms of reference values, the MSE of the restructuring method is only around 17\% lower for $\widetilde{y}_{0.25}$. The difference increases with larger reference values up to more than 25\% for $\widetilde{y}_{0.5}$. 

Retraining has an overall 17\% lower MSE compared to explanation scaling. Over the different datasets, the relative performance advantage of retraining compared to the scaling baseline ranges from 3\% for the boston dataset up to more than 27\% for the friedman dataset.

\begin{figure}[h]
    \centering
    \includegraphics[width=\linewidth]{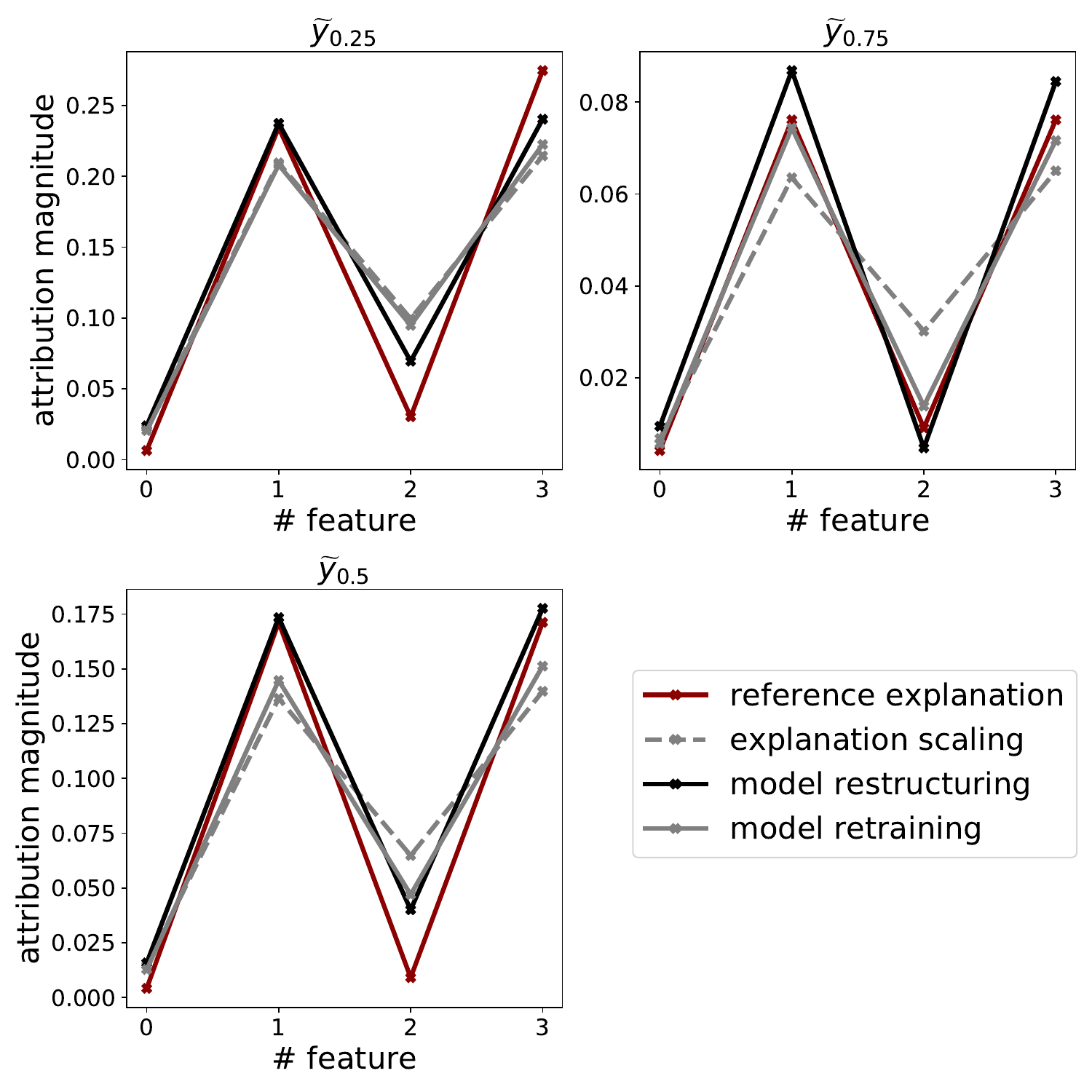}
    \caption{Explanations for a selected example from the friedman dataset, where we compared different explanation methods (explanation scaling, model restructuring and model retraining) to the Shapley-based reference explanation, for the respective reference values $\widetilde{y}_q$.}
    \label{fig:eval_example}
\end{figure}

To gain further insights on the performance improvement of restructuring and retraining vs. explanation scaling we show in Figure \ref{fig:eval_example} an exemplary explanation from the friedman dataset. Firstly, note the different magnitudes on the y-axis induced by the different reference values which all methods are able to account for. However, the restructuring and retraining approach are able to better approximate the structural changes in the attribution with increasing reference values, especially for features of high importance. That illustrates the major benefit of restructuring and retraining over a simple scaling of the original LRP attributions.

Overall, we find the restructuring best suited to explain models with respect to alternative reference values. Furthermore, we see some more options connected to the network structure that could further increase the performance of restructuring. For smaller networks with only 20 neurons, for instance, we have observed the overall MSE difference between restructuring and explanation scaling to increase. This improvement in performance can be explained by the fact that restricted model capacity prevents large components to be scattered onto many neurons and consequently perceived as many small components and flooded. We speculate that approaches that add neurons incrementally during training may also help to further dissociate coarse from fine effects in the architecture and consequently lead to a further increase of explanation accuracy. 

Finally, we recall that in some cases, restructuring may not be possible, for example, because the model's top layer structure does not allow for the proposed adjustments, or may simply not be desirable, because it requires to tamper with the network internals outside the common training and explanation interfaces. In such cases, the more flexible retraining approach (which comes as a close second in our benchmark), can be used as an alternative. Our experiments in the next section will demonstrate the capabilities of the restructuring and retraining approaches on large real-world models.

\section{Application to Real-World Regression Problems}

In this section, we will demonstrate on two real-world application examples, one from the computer vision domain and one from the field of quantum chemistry, the benefits of our XAIR approach. Our approach, which preserves the unit of the prediction in the explanation, and lets the user specify a meaningful reference value, will be shown to enable valuable insights into regression models that cannot be obtained using standard XAI methods out of the box.

\subsection{Facial Image Age Prediction}
As a first illustrative application example, we analyze a popular regression task from computer vision: age prediction from facial images \cite{han2013age, angulu2018age, abdolrashidi2020age}. This problem represents a typical regression task but is often approached by classifying images into different age bins and therefore seems particularly suited to highlight the difference between XAIR and standard XAI. The aim is to highlight the facial features associated with a certain age prediction. This has been done before for the classification approach \cite{lapuschkin2017understanding, abdolrashidi2020age} but, to the best of our knowledge, never for regression models. 

    \begin{figure*}[h]
        \centering
        \includegraphics[width=\linewidth]{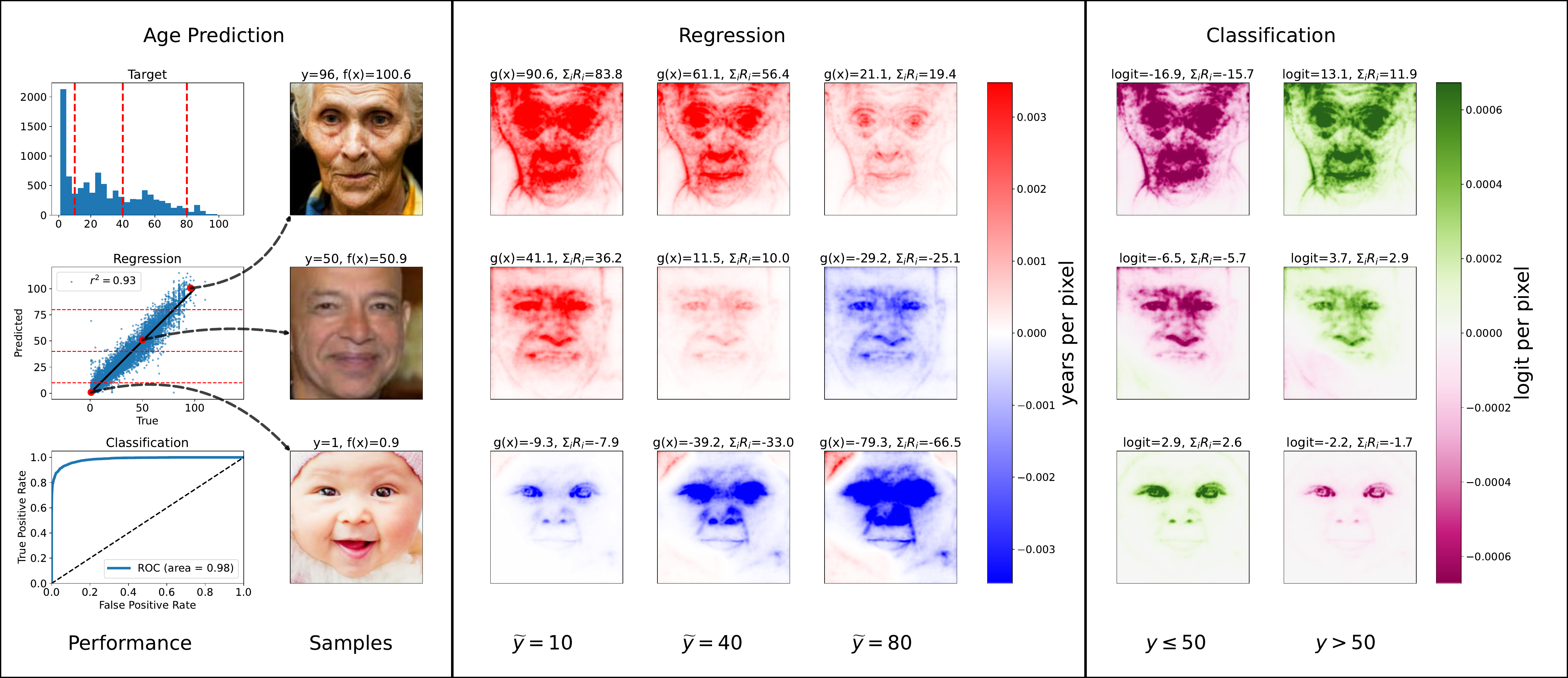}
        \caption{Explanations for age prediction based on facial images. The \textit{Age Prediction} section shows the distribution of targets (top - blue histogram) including the three reference values (10, 40, 80 - vertical red lines), performance plots for both, the regression and the classification model, and the three representative sample images (with true ages 1, 50, 96). The explanations of the regression models for three different reference values are aligned as a 3x3 grid in the center of the figure. Red/blue pixel color indicates positive/negative evidence relative to a given reference point. On the very right vertical section, the classifier decisions are visualized with green/pink pixels indicating evidence for/against the respective class ($>$ 50, $\leq$ 50).}
        \label{fig:age_prediction}
    \end{figure*}   

For this analysis, we make use of a dataset containing $\sim$ 20k facial images associated with biological age\footnote{\url{https://www.kaggle.com/frabbisw/facial-age}} (biased toward younger ages). Each image is pre-processed so that all of them have the same size (200x200) and the faces are aligned and centered. We used a VGG-16 \cite{DBLP:journals/corr/SimonyanZ14a} model pre-trained on ImageNet \cite{DBLP:conf/cvpr/DengDSLL009,DBLP:journals/ijcv/RussakovskyDSKS15} as a feature extractor followed by one ReLU layer with 256 neurons, a dropout-layer, and a final linear layer mapping the 256 neurons to a real-valued age prediction. We selected three representative reference values across the whole target domain, namely 10, 40, and 80 years, and restructured the network's final layers as described in section \ref{sec:restructuring}. To contrast these explanations with those delivered by a standard classification XAI pipeline, we also trained a binary classification model by appending a sigmoid activation layer that classifies whether images are labeled with age below or above 50 years. During training, we minimized the MSE or the binary cross-entropy for the regression and classification cases respectively, both using Adam. In all cases we used \mbox{LRP-$\alpha_1\beta_0$} rule \cite{bach-plos15,montavon2018methods} in the convolutional layers and \mbox{LRP-$\epsilon$} rule \cite{bach-plos15} (where biases are ignored) for the fully connected layers.

Because retraining is very time consuming in case of deep and complex models, we opt for the restructuring approach. For restructuring the model with respect to the desired reference value we used the flooding approach described in Section \ref{sec:restructuring}. For the cases where $f(\x) < \widetilde{y}$, the flooding equation was slightly modified to $(\ba - (t\cdot\boldsymbol{1}_{w\leq0}-t/4\cdot\boldsymbol{1}_{w>0}))^+ , t \in \mathbb{R}$, where the asymmetry between positive and negative weights allows to explore activation patterns associated to higher ages.

The results of this experiment are presented in Fig.\ \ref{fig:age_prediction}, where we picked three representative examples from the test set such that the actual age is in relative proximity to a selected reference value. Looking at the regression explanations (center block of the figure) we observe a gradient from strongly red heatmaps (composed of positive scores) in the top-left corner, to strongly blue heatmaps (composed of negative scores) in the bottom-right corner. This reflects the conservative nature of the explanation technique where the sum explanation scores correspond approximately to the signed difference between the predicted age $f(\x)$ and the chosen reference value $\widetilde{y}$.

When looking at the explanations for the 96 year old person (top row) with respect to reference value of age 10 (left) we observe that all facial features significantly contribute to the high age model output. The closer the reference point moves to the true age, the more differentiated the heatmap becomes. It becomes apparent that it is mainly the area around the person's eyes that, according to the model, makes the person look older than, for example, 80 years (right). This example once more emphasizes one of the essential motivations for this paper: In regression, explanations relative to faraway reference values might be useful in certain cases, but often it is more interesting to explain relative to a close-by reference value.  

Further insights about the model can be gained from the 50-year-old person's explanations. Looking at the explanation relative to the nearest age reference $\widetilde{y}=40$ (in the very center of the figure), we observe an absence of strongly positive or negative scores over the person's facial features. This indicates that, according to the model, the person does not have any facial features that would be particularly atypical for his age group. In comparison, classifier-based explanations could not convey such fine-grained insights.

Finally, the baby picture explanations (bottom row) allow for similar conclusions as for the old person's example (but with a flipped sign). The explanation relative to reference value age 10 (left) shows that the model mainly used the region around the eyes to decide that the person was younger than 10 years and explanations for reference points further away from the true age do not allow for such a detailed interpretation. Moreover, the example enables another interesting observation: with rising reference values the model allocates more importance to the babies hairline/hat-area. A possible interpretation is that in the context of young age headware is irrelevant as the model mainly uses the young facial features for its judgement. In reference to older ages, however, the model seems to have associated the hat with an age increase, which could be due to the hat resembling white hair. This illustrates once more how the choice of different reference values allows for a contextualization of the explanations that reaches beyond the abilities of any classification model (compare classification heatmaps on the right side). 

In summary, the experiment shows that our XAIR approach allows for richer explanations and therefore more insights into what features make a face appear to belong to a certain age prediction. Moreover, the preservation of units allows for an attribution of a number of years to specific facial features, a quality that might be even more beneficial in other domains, such as quantum chemistry (cf.\ Section \ref{sec:quantum}).

\subsection{Explanations in Quantum Chemistry}
\label{sec:quantum}

Recently, ML methods have contributed broadly to atomistic simulations e.g.\ for ultra-fast approximate solutions of the Schroedinger Equation (factor $10^7$ faster \cite{rupp2012fast,von2020exploring}) and fast and accurate force fields for molecular dynamics (e.g.\ \cite{chmiela2017machine,noe2020machine,unke2020machine}). Especially, the development of ML architectures for predicting molecular properties has been advancing rapidly with ever more complex models achieving gradual improvements in performance (e.g.\ \cite{rupp2012fast, schutt2017quantum, gilmer2017neural, schutt2018schnet, physnet, klicpera2020directional}). However, a good model should exhibit qualities beyond high accuracy, namely, it should be able to capture certain principles from physics \cite{schutt2017quantum,von2020exploring}. This allows for increased confidence in the model results through sanity-checks by an expert as well as new insights into physical phenomena previously not understood (e.g.\ \cite{schutt2017quantum,keith2021combining,unke2021spookynet}). XAI methods, especially XAIR methods, can be a key for both which we will demonstrate in the following sections by applying our proposed retraining approach in the quantum chemistry domain.

In particular, we have a closer look at the effect of different reference values when explaining predicted molecular quantum-chemical properties. These properties are real-valued and expressed in various physical units (e.g.\ kcal/mol, etc.). We chose to predict the atomization energy, which is the energy of a molecule relative to a total separation into individual atoms. The atomization energy is a negative quantity, since forming a molecule is energetically preferable compared with a separation into individual atoms. Another way of thinking about the atomization energy is that it is the energy needed for breaking all bonds of the compound and separating its atoms infinitely far apart, and which is effectively equivalent to the `negative' atomization energy.

For the prediction of the negative atomization energy, which we will simply call ``energy'' from now on, we utilize the SchNet~\cite{schutt2017schnet, schutt2018schnet, schutt2018schnetpack}, a GNN architecture specifically designed for predicting molecular properties. For the explanations, we employ a higher-order extension of LRP designed for explaining GNNs (GNN-LRP \cite{schnake2020xai}). The output prediction is attributed to so-called walks that comprise several nodes and edges of the graph. Since the predicted energy corresponds to interactions between atoms, we expect most relevance to be attributed to the edges and in particular to the bonds. Furthermore, it is common sense that bonds of higher order are more stable and thus more energy is needed to break them and it is important that SchNet models also capture this property. We will show that using a reasonable reference value facilitates the extraction of insights from the model, by precisely answering the following questions: 
    \begin{enumerate}
        \item Are bonds accountable for the major energy contributions (as chemists would expect)?
        \item Are the energy contributions of bonds increasing with increasing bond order (as chemists would again expect)?
    \end{enumerate}

\begin{figure}[h]
        \centering
        \includegraphics[width=0.49\textwidth]{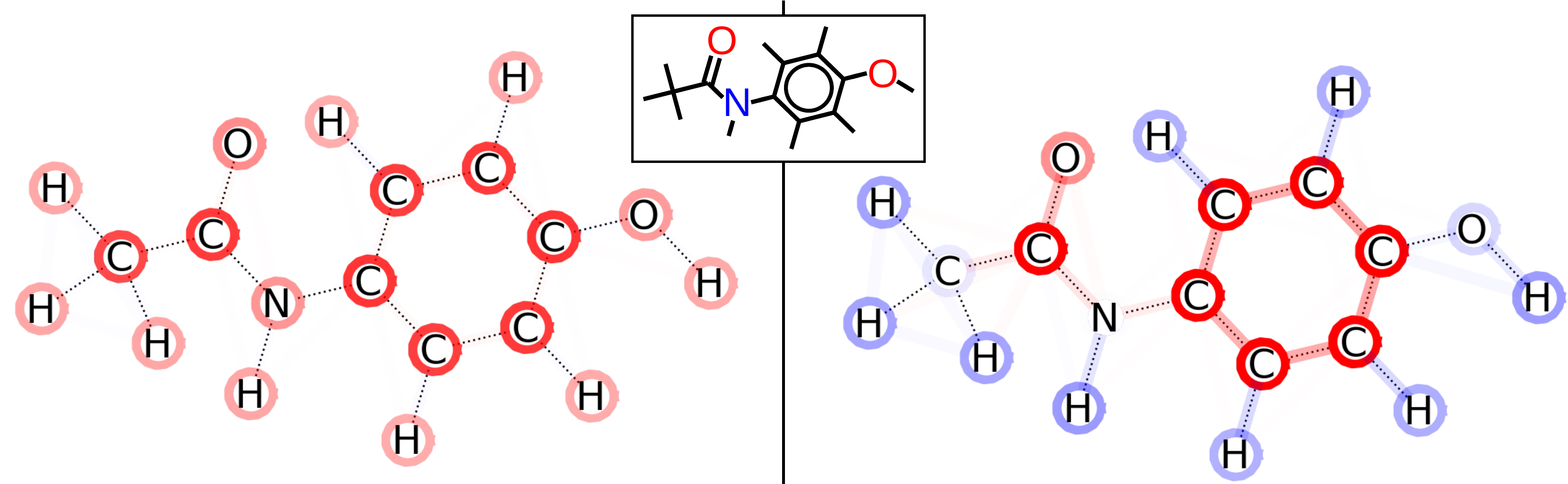}
        \caption{Comparison between (left) SchNet trained with zero reference value, and (right) SchNet retrained with non-zero reference value for paracetamol (acetaminophen). The heatmaps show the aggregated relevance scores for atoms and bonds, respectively. The structure formula shows the bond orders present in the molecule.}
        \label{fig:comp_ref_value}
    \end{figure}

In our experiments, we first train the SchNet model with reference value zero (achieving MAE of $\SI{0.017}{\electronvolt}$) and compute the respective explanations. To get more contextualized explanations, we consider a reference value corresponding to the mean energy per atom, averaged over all molecules consisting of single bonds only, and apply our retraining approach. For the retrained model (MAE of $\SI{0.015}{\electronvolt}$), the average relevance attribution of single atoms should be driven close to zero, and we should therefore be able to better distinguish the different bond orders in the explanation. In Fig.\ \ref{fig:comp_ref_value}, we compare the explanation of the two SchNet models mentioned above. The explanation (relevance heatmap) corresponding to the SchNet trained with reference value zero is depicted in Fig.\ \ref{fig:comp_ref_value}~(left). The strength of different bonds can hardly be distinguished, and a large part of the relevance is attributed to the atoms and not to the bonds. In contrast, for the model trained with non-zero reference value shown in Fig.\ \ref{fig:comp_ref_value}~(right), the bond structure becomes more prominent, and we can also distinguish between bonds of different bond strengths. The double bond and the aromatic ring exhibit large relevance scores, while little relevance is attributed to single bonds.
    
\begin{figure}[h]
    \centering
    \includegraphics[width=0.34\textwidth]{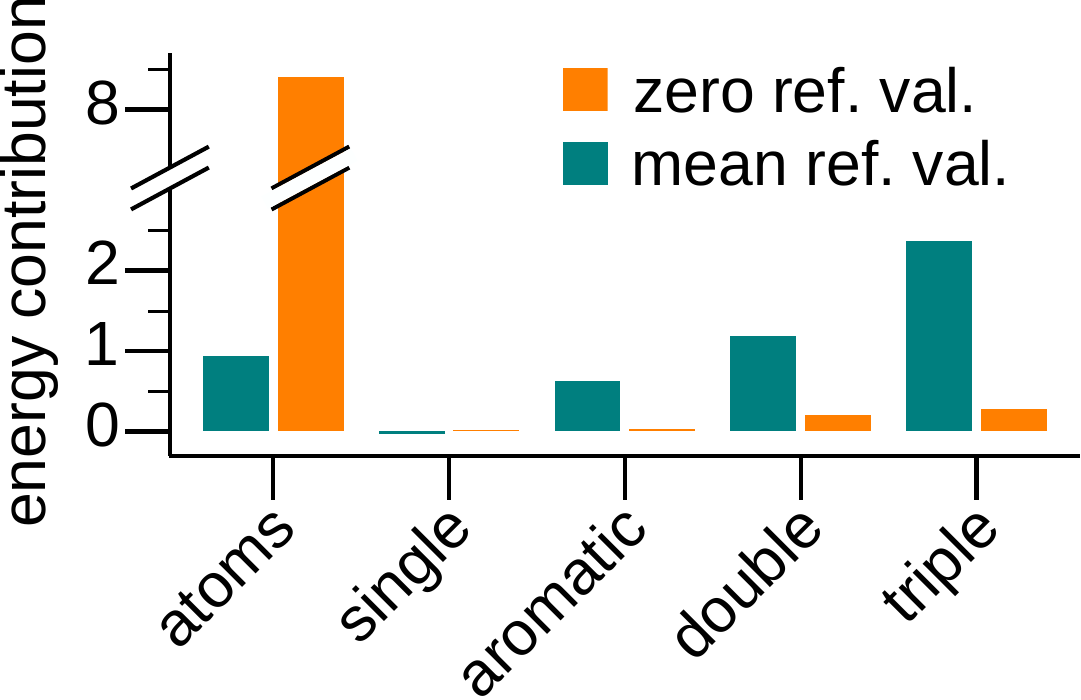}
    \caption{Average energy contribution per bond and atom for a SchNet model trained with zero-reference value (zero ref.\ val.) and another SchNet model trained with non-zero-reference value (mean ref.\ val.). The bonds are distinguished by bond order. The depicted energy contribution has arbitrary units.}
    \label{fig:at_bond_stat_comp}
\end{figure}

To substantiate the observations above, we evaluate the relevance attributions of the two models for a set of 100 molecules randomly drawn from the QM9 dataset \cite{rupp2012fast}. To this end, we calculate the mean relevance attributions corresponding to atoms and different bond types for the two SchNet models, respectively. Relying on the relevance conservation, we can associate the relevance aggregated on particular bonds and atoms as their respective energy contributions, as done in \cite{schnake2020xai}. The evaluation is shown in Fig.\ \ref{fig:at_bond_stat_comp}. Both models capture the physics w.r.t.\ bond orders fairly well, which is an increasing energy contribution with increasing bond order. For the SchNet model trained with reference value at zero, the atoms contribute the most energy to the total prediction, while the atoms of the model trained with non-zero reference contribute less. This explains why even though both models are designed according to chemical intuition, the predictions of the zero reference model are hardly interpretable. It is because the large relevance attribution to atoms dominates the relevance attribution of interest on the bonds.

To further emphasize the importance of choosing a well-suited reference value for explaining models from quantum chemistry, we propose the following experiment: The QM9 dataset consists of relaxed molecules in their ground state. Distorting the structure of these molecules would hence yield energetically less favourable states. It is essential that the ML model captures this feature by predicting an energy change, in our case to lower energies. To this end, we consider a H\textsubscript{7}C\textsubscript{5}N\textsubscript{3}O molecule from the QM9 dataset. We predict the energy for the molecule in its relaxed state as well as for a distorted structure, where we stretched the bond between the Carbon and the Oxygen atom by $\SI{0.4}{\angstrom}$. The zero reference model and the non-zero reference model predict an energy change of $\SI{1.45}{\electronvolt}$ and $\SI{2.46}{\electronvolt}$, respectively. Even though the model predictions for the distorted molecule differ from each other, both models predict a decrease in energy, which is what we expect from a well-trained model. For both models, we compute the relevance scores for the ground state structure and for the distorted structure, respectively. A reasonable relevance attribution is expected to indicate the region on the molecule that is responsible for the energy change.
    
\begin{figure}[h]
    \centering
    \includegraphics[width=.75\linewidth]{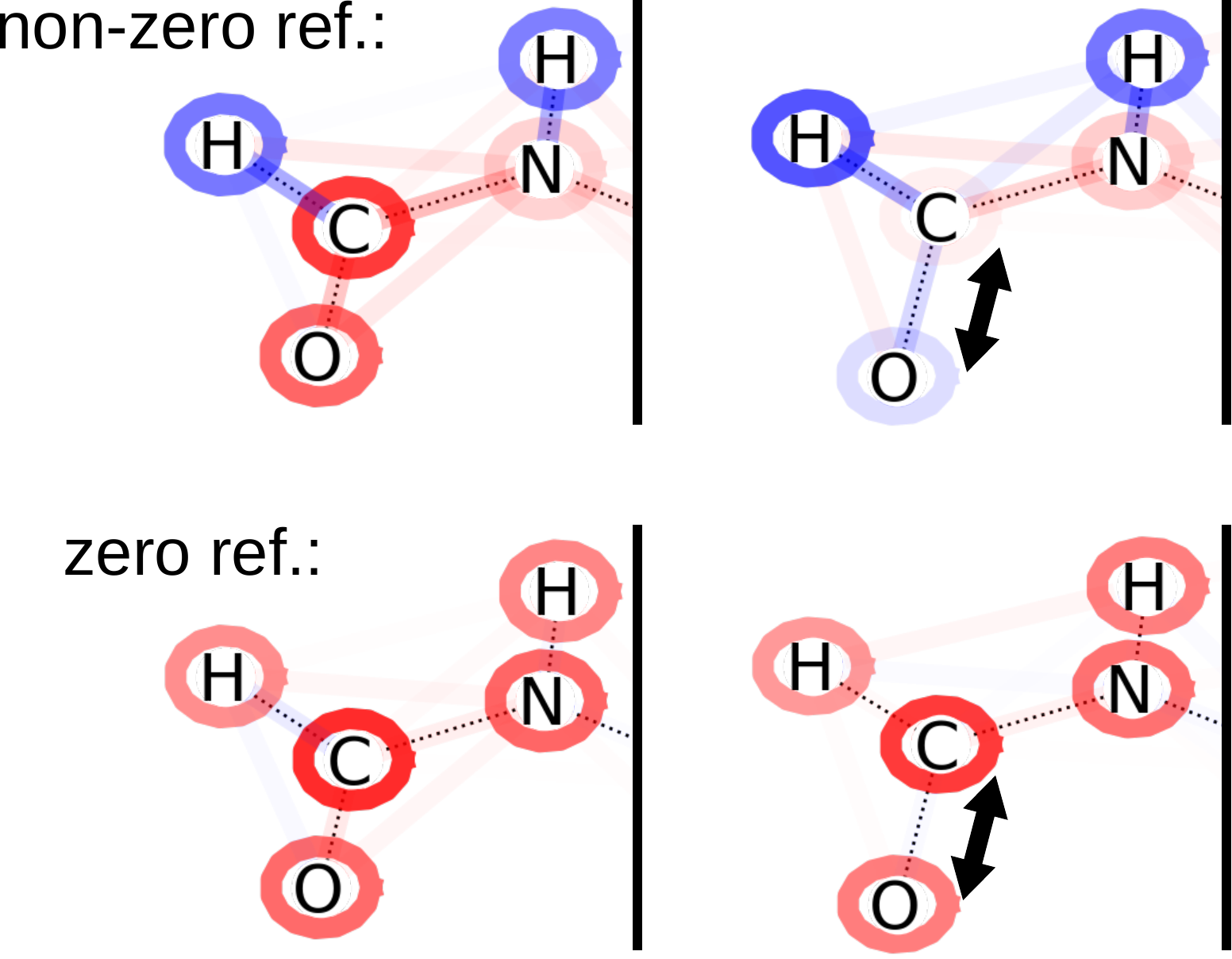}
    \caption{Relevance scores obtained by GNN-LRP for the (left) relaxed and (right) distorted geometry of H\textsubscript{7}C\textsubscript{5}N\textsubscript{3}O. The distorted molecule is obtained by stretching the C-O bond as indicated by the arrow. The first row shows the relevance attribution for the non-zero reference model. The second row shows the relevance heat-map for the zero reference model. The depicted relevance scores are aggregated for each bond and atom, respectively. To increase visibility of the edge relevance attribution, we show the scaled relevance scores $R' = R^{0.7}$.}
    \label{fig:distorted}
\end{figure}

The results are depicted in Fig.\ \ref{fig:distorted}. For the non-zero reference model, we can clearly see a local change in the relevance heatmap when stretching the C-O bond. The edge and atom relevance in the distorted region strongly decreases and even switches to a negative value. Comparing left and right heatmap shows that the structure distortion results in a less favourable molecule conformation. Furthermore, the heatmap clearly indicates which substructure of the molecule is responsible for the energy difference with respect to the relaxed structure. For the zero reference model, in contrast, we can hardly spot any difference between the heatmaps of relaxed and distorted structures.
    
In conclusion, both SchNet models solve the regression task with comparable accuracy. The statistical evaluation shows that both models capture the physics behind the problem. However, choosing a well-suited reference value clearly facilitates explaining the distinction between physical concepts known to an expert, such as bond orders. Moreover, it enables the identification of energetically (un)favorable substructures. It thereby offers a better validation of the model to a chemist user, in line with chemical intuition.

\section{Best practice guidelines}

This section provides a set of good practices in order to successfully produce XAIR explanations. The discussion focuses on the two main aspects treated in this paper, which are (1) the conservation of measurement units from the modeled system up until the production of the explanation, and (2) the incorporation of the reference value specified by the user into the explanation process. Our recommendations are meant to apply regardless of the concrete XAIR application, or the exact choice of neural network model. For more general recommendations on XAI we refer to \cite{DBLP:journals/pieee/SamekMLAM21}. 

\subsection{Conservation of measurement units}
\label{sec:best_practice_pu}

Many ML tasks have the objective to predict a quantity that inherently is expressed with some measurement unit, whether it is the energy of a physical system, the value of an economic good, or else. Often, for the purpose of facilitating training, the targets are transformed in some way, such as converting the real-valued prediction problem into a simpler problem (e.g.\ classification or ranking), or applying some nonlinear transformation. When the measurement units are of interest, we discourage such intermediate step as it loses the latter interpretation.
\begin{center}
\medskip
\fbox{\textbf{Predict using the original measurement unit.}}
\medskip
\end{center}
To ensure that the lack of target normalization does not make training more difficult, we can still perform the training in transformed space (e.g.\ standardized targets) but we need to make sure the model is rescaled back to the original units after training.

We now would like to translate the unit-preserving predictions into unit-preserving explanations. Explanation techniques such as the Shapley value, integrated gradients, or LRP are all based on a conservation principle, which when applied to a real-valued prediction, enables an understanding of the produced explanation in terms of the same measurement unit.
\begin{center}
\medskip
\fbox{\textbf{Use a conservative explanation technique.}}
\medskip
\end{center}
By applying this recommendation, we ensure that such a precise meaning of the explanation is available, and it can be used to make quantified inferences, which are more useful to the user than a simple visualization. Examples include estimating the stability of some subgroup of atoms in a molecule, or estimating the value of individual components of some more complex economic good.

\subsection{Choosing a reference value}

The choice of reference value has a profound impact on the quality of the explanations as shown multiple times throughout the paper. This being said, it is impossible to recommend a general reference value due to the multitude of potential questions the practitioner might want to ask using the XAI method. Therefore the appropriate choice relies solely on the respective applicant's question and domain knowledge.
\begin{center}
\medskip
\fbox{\textbf{Ask the user for an appropriate reference value $\widetilde{y}$.}}
\medskip
\end{center}
Once a reference value has been specified by the user, one needs to ask how to integrate this information into a meaningful explanation. Various ways of producing such an explanation have been described in Section \ref{sec:method}, in particular, we have proposed \textit{retraining} and \textit{restructuring} as two practical solutions that enable scaling to large models. In general, if restructuring is possible, i.e.\ the top-level structure of the neural network consists of a ReLU layer followed by some linear layer (this includes simple average pooling), we recommend using this approach as it shows the highest performance in our benchmark while requiring least intervention. 
\begin{center}
\medskip
\fbox{\textbf{If possible, incorporate $\widetilde{y}$ using restructuring.}}
\medskip
\end{center}
If instead the structure of the network is different, or even if the structure is appropriate but we do not want to tamper with the neural network internals and use exclusively the usual training and explanation interfaces, the retraining approach becomes the favorable alternative, provided that we have access to the training data and some compute resources. Retraining is however a more complex procedure that requires further heuristics. We address this topic in the section below.

\subsection{Retraining guidelines}
\label{section:retraining-guidelines}
Apart from being able to manage potentially high associated computational cost, the most important practical requirement for retraining is to have access to a suitable training dataset. In practice this might be prevented for legal reasons such as privacy or close source as well as prohibitively large dataset size. For the data to be suitable it is crucial to have the same (or similar) distribution as the original data to arrive at a model which implements $g(x)$ as precisely as possible.
\begin{center}
\medskip
\fbox{\textbf{Ensure access to appropriate training data.}}
\medskip
\end{center}
Assuming we have the necessary data, and once the corresponding targets are adjusted to the desired reference value, the model implementing $g(x)$ can be retrained, if possible, using the original training configuration. For coherence, we recommend training with the adjusted target at least until the training error reaches a comparable level as in the original model training, rather than training for a fixed number of epochs.
\begin{center}
\medskip
\fbox{\textbf{Retrain until the original accuracy is reached.}}
\medskip
\end{center}
Because the retraining might substantially perturb the original learned solution, it might be preferable to freeze certain parts of the original model such as the low-level layers extracting generic features, and retraining only the top layers. Alternatively, in order to not tamper with the existing features, all weights of the model can be frozen, and only the biases attached to features at each layer are adjusted. Importantly we also need to avoid the scenario where the top-layer bias would simply shift to the new reference value thereby preventing any meaningful change in the explanation. If this happens, we recommend freezing the top-layer biases to their original value and only learning biases in the layers below.
\begin{center}
\medskip
\fbox{\textbf{Constrain the parameters to avoid trivial solutions.}}
\medskip
\end{center}
In any case, we recommend to closely monitor model retraining to ensure that the imposed constraints allow for a satisfactory solution of the adjusted regression problem. A further degree of freedom is the selection of the training data. If interested in a particular reference value, we recommend excluding training data points with target values too far from the selected reference value $\widetilde{y}$, by choosing appropriate band parameters $\tau^+$ and $\tau^-$. This last recommendation however holds only to the extent that generalization performance of the model is not penalized by such exclusion. Lastly, model training and retraining is a complex process with many degrees of freedom and tools for analyzing success. Here, general guidance on training neural networks or other nonlinear models remains applicable (e.g.\ \cite{GoodBengCour16,intronn,braun2008relevant,montavon2011kernel,lecun2012efficient}).

\subsection{Implementation and sanity checks}

In practice, because the methods we propose for explaining regression models do not require a change of the explanation technique but only a modification of the function or of the architecture implementing such function, existing XAI frameworks remain applicable. Software packages, such as Captum\footnote{\url{https://captum.ai/}}, iNNvestigate \cite{DBLP:journals/jmlr/AlberLSHSMSMDK19}\footnote{\url{https://github.com/albermax/innvestigate}}, DeepExplain\footnote{\url{https://github.com/marcoancona/DeepExplain}} and Zennit \cite{anders2021software}\footnote{\url{https://github.com/chr5tphr/zennit}} implement a number of popular explanation methods. Some of them, such as iNNvestigate, come with an implementation of LRP. We provide an implementation of our restructuring approach \footnote{\url{https://github.com/sltzgs/xai-regression}} which enables a seamless integration into existing XAI pipelines and frameworks.

\smallskip

Whether the practitioners use a third party implementation or their own implementation, there are a number of sanity checks (or unit tests) that can be applied to verify that the procedure satisfies some basic necessary conditions. The most important one is \textit{conservation}, in particular, on needs to verify that $\sum_i R_i = f(\x) - \widetilde{y}$. If the network has biases and one can consequently not expect the explanation to be fully conservative, the sanity check can still be applied on a version of the network where the biases are turned to zero. Other sanity checks are specific to the method. For example, one can verify that LRP explanations reduce to special (and easy to implement) cases, for particular choices of LRP hyperparameters. For example, we can verify that an implementation consisting of LRP-$\epsilon$ and LRP-$\gamma$ rules reduces to Gradient$\,\times\,$Input if setting the hyperparameters $\epsilon$ and $\gamma$ to zero.

\section{Conclusions \& Outlook}

Explainable AI has demonstrated great potential in making the decision of black-box models transparent to the user. Recently, methods have been proposed to scale explanations to highly complex neural network classifiers composed of millions of neurons. ML is, however, a vast field, with many formulations of the learning problem that address the numerous application scenarios encountered in practice, including unsupervised learning, supervised learning, reinforcement learning, and within supervised learning, subcategories such as classification or regression or ranking.

In this paper, we found that explanation methods that are based on a conservation principle (e.g.\ Shapley values, integrated gradients, or LRP) are particularly favorable in the regression scenario as they allow for a decomposition of the predicted quantity on the input features that preserves an interpretation in the same measurement units as the prediction tasks (e.g.\ units of energy or monetary units). However, we have also demonstrated that Explainable AI cannot simply be transferred between different types of ML problems without adaptation. In particular, an out-of-the-box deployment of explanation techniques for classification omits the fact that a real-valued prediction must often be explained with respect to a particular reference value for the explanation to be meaningful.

We have proposed a collection of techniques for incorporating such reference values, that addresses the various constraints of the explanation method or the model used for prediction. In particular, we have contributed a lightweight restructuring approach, that enables an accurate incorporation of the reference value into the explanation without having to retrain the model nor having to evaluate the function multiple times. This restructuring approach is conceptually related to the `neuralization' approach \cite{DBLP:journals/corr/abs-1906-07633,DBLP:journals/pr/KauffmannMM20} which was proposed to extend XAI methods from supervised to unsupervised learning, but this time it was used for extending from classification to regression.

After verifying the performance of our approaches on a set of benchmark experiments, we have demonstrated concrete practical use cases on image data as well as on molecular data from atomistic simulations, where chemically plausible explanation of molecular energy could be produced.

\medskip

As our paper contributes a first step toward extending XAI to regression in a systematic and theoretically founded manner, further work is needed to address other specificities of the regression task over simple classification, such as model uncertainty which is typically decoupled in the regression case from the prediction itself. Ways to combine the prediction and the uncertainty of the prediction into a single explanation remains an open question.

Furthermore, while we have demonstrated empirically a close correspondence between our explanations and a putative reference explanation based on the Shapley value, thereby demonstrating the faithfulness of our method, it will be necessary to extend the evaluation to account for the overall benefit to the user. 
This could take the form of user experiments, which typically take into consideration whether an attribution on input features is helpful to the user, or whether more structured explanations would be desirable. Such explanations could for example be based on extracted interpretable latent variables, which would allow to not only answer \textit{what} input features are important, but also \textit{why} they are important for a given prediction.

Finally, to further assess the overall practical benefits of using explanations, it will be important to consider end-to-end scenarios \cite{DBLP:journals/corr/Doshi-VelezK17}, where one can precisely measure the utility gain of incorporating explanations into a given learning system, over not incorporating them, in particular, how much a user can objectively gain from using an explanation expressed in the same measurement units as the prediction, and from the additional level of granularity offered by choosing a particular reference value. Also related to the question of evaluation is the level of degradation that can be expected if allowing for adversarial entities into the prediction or explanation process \cite{DBLP:conf/nips/DombrowskiAAAMK19}, and what countermeasure can be provided to such scenario. Lastly, manageability and energy efficiency aspects such as the cost of implementing, maintaining, and running XAI approaches in an environment of increasingly complex ML models, should also be taken into account in the overall assessment. 

\section*{Acknowledgements} 
This work was partly funded by the German Ministry for Education and Research (under refs 01IS14013A-E, 01GQ1115, 01GQ0850, 01IS18056A, 01IS18025A and 01IS18037A), the German Research Foundation (DFG) as Math+: Berlin Mathematics Research Center (EXC 2046/1, project-ID: 390685689), the Investitionsbank Berlin under contract No.\ 10174498 (Pro FIT programme), and the European Union’s Horizon 2020 research and innovation programme under grant agreement No.~965221.
Furthermore KRM was partly supported by the Institute of Information \& Communications Technology Planning \& Evaluation (IITP) grants funded by the Korea Government (No. 2019-0-00079, Artificial Intelligence Graduate School Program, Korea University). 
Correspondence to WS, KRM and GM.

\bibliographystyle{abbrv}
\bibliography{bibliography}

\end{document}